%% file: main.tex
\documentclass[10pt,twocolumn,letterpaper]{article}

\usepackage[pagenumbers]{cvpr} 

\usepackage{xcolor}
\usepackage{colortbl}
\definecolor{dcolor}{RGB}{204,229,255}
\definecolor{cyan1}{RGB}{0,255,255}

\usepackage{wrapfig}
\usepackage{makecell}
\usepackage{multirow} 

\usepackage{etoolbox}
\makeatletter
\robustify\@latex@warning@no@line
\makeatother
\usepackage{authblk}


\definecolor{cvprblue}{rgb}{0.21,0.49,0.74}
\usepackage[pagebackref,breaklinks,colorlinks,allcolors=cvprblue]{hyperref}

\def\paperID{1719} 
\def\confName{CVPR}
\def\confYear{2025}

\title{Seeing Beyond Views: Multi-View Driving Scene Video Generation \\ with Holistic Attention}

\author{
Hannan Lu\textsuperscript{1} \quad 
Xiaohe Wu\textsuperscript{1} \quad
Shudong Wang\textsuperscript{2} \quad
Xiameng Qin\textsuperscript{2} \quad
Xinyu Zhang\textsuperscript{3} \newline
Junyu Han\textsuperscript{2} \quad 
Wangmeng Zuo\textsuperscript{1} \quad 
Ji Tao\textsuperscript{2} \quad
\\
\textsuperscript{1}Harbin Institute of Technology \quad
\textsuperscript{2}Changan Automobile \quad 
\textsuperscript{3}The University of Adelaide \quad 
}

\begin{document}
\maketitle

\input{sec/0_abstract}    
\input{sec/1_intro}

\input{sec/2_related_work}
\input{sec/3_method}

\input{sec/4_experiment}

\input{sec/5_conclusion}
{
    \small
    \bibliographystyle{ieeenat_fullname}
    \bibliography{main}
}
\input{sec/X_suppl}


\end{document}

%% file: sec/0_abstract.tex
\begin{abstract}

Generating multi-view videos for autonomous driving training has recently gained much attention, with the challenge of addressing both cross-view and cross-frame consistency.
Existing methods typically apply decoupled attention mechanisms for spatial, temporal, and view dimensions.
However, these approaches often struggle to maintain consistency across dimensions, particularly when handling fast-moving objects that appear at different times and viewpoints.
In this paper, we present CogDriving, a novel network designed for synthesizing high-quality multi-view driving videos. 
CogDriving leverages a Diffusion Transformer architecture with holistic-4D attention modules, enabling simultaneous associations across the spatial, temporal, and viewpoint dimensions.
We also propose a lightweight controller tailored for CogDriving, \ie, Micro-Controller, which uses only 1.1\% of the parameters of the standard ControlNet, enabling precise control over Bird’s-Eye-View layouts.
To enhance the generation of object instances crucial for autonomous driving, we propose a re-weighted learning objective, dynamically adjusting the learning weights for object instances during training.
CogDriving demonstrates strong performance on the nuScenes validation set, achieving an FVD score of 37.8, highlighting its ability to generate realistic driving videos.
The project can be found at \href{https://luhannan.github.io/CogDrivingPage/}{https://luhannan.github.io/CogDrivingPage/}.

\end{abstract}


%% file: sec/1_intro.tex
\section{Introduction}
\label{sec:intro}

\begin{figure}[h]
    \centering
    \includegraphics[width=0.9\linewidth]{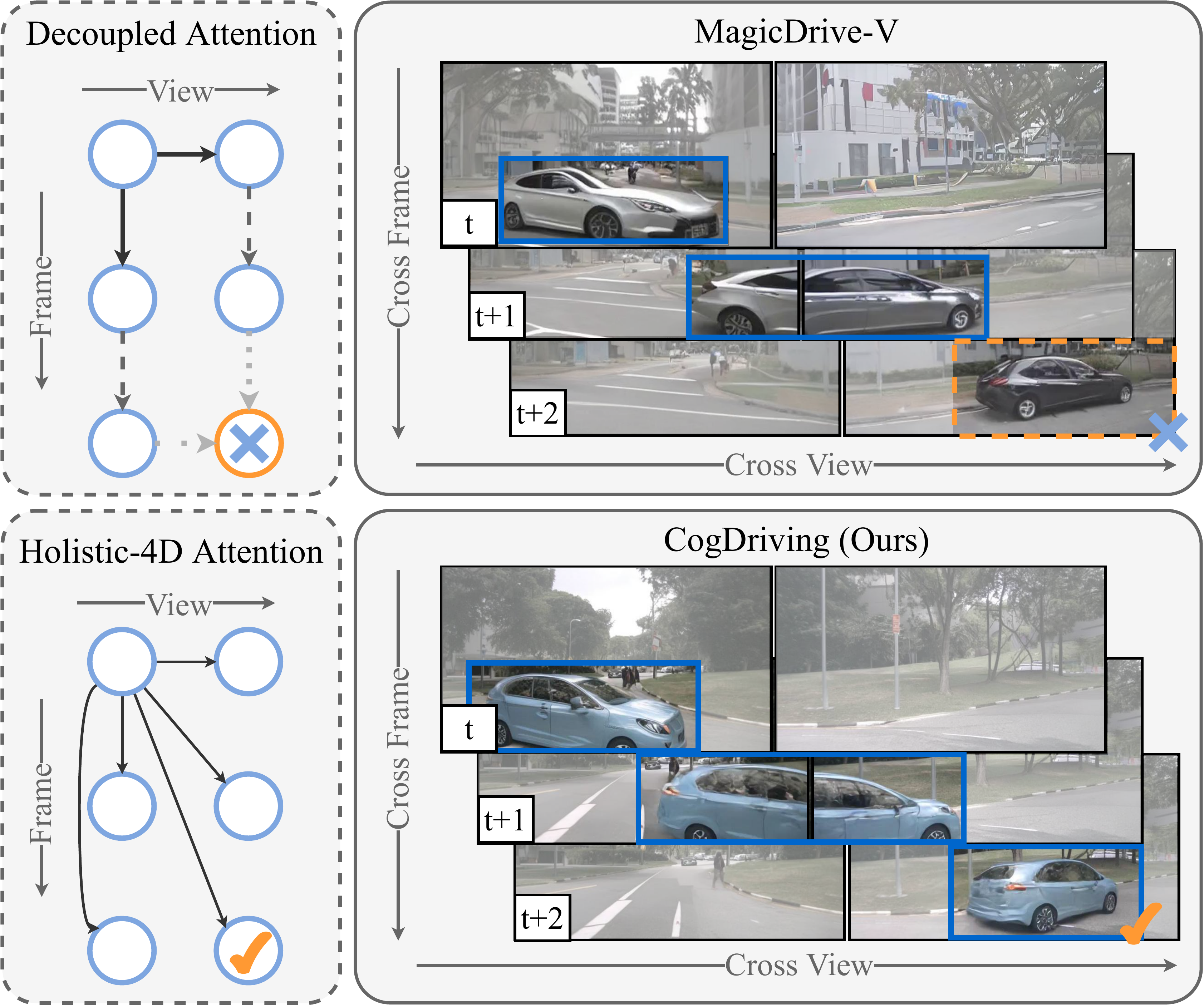}
    \vspace{-2mm}
    \caption{Decoupled Attention~\cite{gao2024magicdrive3d} vs Holistic-4D Attention.
    Our Holistic-4D attention establishes straightforward cross-dimensional relationships, leading to explicit transmission of visual information and enhanced cross-view consistency.
    }
    \label{fig:fig2}
\end{figure}

With the advancements in autonomous driving, Bird's Eye View (BEV) perception techniques have become fundamental and achieved significant improvement in critical tasks such as 3D object detection\cite{huang2021bevdet, li2022bevformer}, map segmentation~\cite{jiang2023polarformer, liu2023petrv2}, multi-object tracking~\cite{hu2022monocular, zhang2022mutr3d, zhang2303bytetrackv2}, and 3D lane detection~\cite{huang2023anchor3dlane, chen2022persformer}. 
The success of BEV perception techniques heavily relies on learning from multi-view videos.
However, the collection and annotation of such data remain labor-intensive and challenging, especially in extreme weather conditions and rare scenarios.
To resolve this challenge, recent methods~\cite{yang2023bevcontrol, gao2023magicdrive, gao2024magicdrive3d, li2023drivingdiffusion, wen2024panacea} have increasingly turned to generative models, particularly diffusion-based models, to synthesize multi-view driving videos conditioned on BEV layouts.

In contrast to conventional video generation models that focus primarily on maintaining temporal coherence within individual videos, \ie, cross-frame consistency, multi-view video generation requires addressing both cross-frame and cross-view consistency simultaneously.
The cross-view consistency refers to ensuring that the content captured across different camera perspectives within a given scene remains consistent.
To this end, recent diffusion-based multi-view video generation methods~\cite{wen2024panacea, li2023drivingdiffusion, gao2024magicdrive3d} typically decouple the view dimension from the spatial-temporal dimensions and incorporate additional attention modules to model dependencies across different camera views.
In practice, the consistency of cross-frame and cross-view is coupled together as shown in Fig.~\ref{fig:fig2}.
For example, when a vehicle moves from the Front-Left view at the $t$-th frame to the Front-Right view at the $(t+2)$-th frame, the method with decoupled attention~\cite{gao2024magicdrive3d} fails to maintain a consistent appearance between frames.
This limitation arises from the decoupling of attention modules, which leads to extensive implicit transmission for cross-dimensional associations. 
This significantly increases the learning complexity, making it challenging to maintain the consistency of content across dimensions. 
Therefore, \textbf{seeing beyond views} is crucial to multi-view consistency across frames, and a method capable of establishing holistic relationships between spatial, temporal, and viewpoint dimensions is essential for generating high-quality multi-view videos.

In this work, we present \textit{CogDriving}, a novel network designed for synthesizing consistent multi-view driving videos. 
CogDriving introduces a simple yet effective diffusion transformer equipped with holistic-4D attention module to simultaneously model associations across spatial, temporal, and view dimensions.
By establishing complete and straightforward cross-dimensional relationships, it significantly simplifies the learning process and enhances the ability to generate consistent content across all dimensions.
As illustrated in Fig.~\ref{fig:fig2}, the direct cross-dimensional associations established through the holistic-4D attention mechanism in CogDriving ensure the consistency of the vehicle's appearance across different frames and views.

Alongside consistency, enabling controllability is a fundamental requirement in multi-view driving scene video synthesis.
Controllability ensures that the generated multi-view content conforms to the BEV layouts that characterize the scene (road maps, 3D bounding boxes, camera poses, etc.). 
Prevailing methods~\cite{yang2023bevcontrol, gao2023magicdrive, li2023drivingdiffusion, wen2024panacea} typically utilize ControlNet~\cite{zhang2023adding}, a widely adopted paradigm for conditional generation that duplicates the diffusion U-Net to process conditional information.
In our case, duplicating the network with 4D attention to process conditional inputs is impractical, as it introduces excessive computational overhead and hinders network optimization and convergence.
%
%
Inspired by the lightweight encoder design and cross-normalization mechanism introduced in ControlNeXt~\cite{peng2024controlnext}, we propose the Micro-Controller tailored to the multi-view generation tasks for the integration of conditioning controls.
Specifically, to handle the various types of input conditions in driving video generation, we employ separate encoders and independent cross-normalization for each type of condition.
Additionally, to achieve spatial-temporal alignment between each condition embedding and the multi-view latents, we deploy an encoder architecture with 3D convolution and temporal down-sampling layers.

In street-view videos, object instances, such as cars and pedestrians, typically occupy a smaller spatial area relative to the background, such as roads, \etc.
This imbalance in spatial distribution leads the model to prioritize learning the background content generation rather than rendering object instances, which are key elements in autonomous driving systems.
To enable the model to balance the generation of object instances and background content, we propose a re-weighted learning objective to dynamically assign learning importance to object instances during training.
Without increasing computational overhead during inference, such a learning objective effectively enhances the capability of instance generation.

Our main contributions can be summarized as follows:

\begin{itemize}

\item 
We propose CogDriving, an innovative DiT equipped with holistic-4D attention module for multi-view video generation, which simultaneously models associations across spatial, temporal, and view dimensions.

\item 
We introduce a lightweight control branch for controllable generation, \ie, Micro-Controller, which is tailored to our DiT with 4D attention, and contains only 1.1\% of the parameters of the ControlNet branch, yet achieving competitive control over generated results.

\item 
We devise a re-weighted learning objective that emphasizes the supervised learning of object instances. 

\item 
CogDriving attains an FVD score of 37.8 on the nuScenes dataset, highlighting its proficiency in generating high-quality driving-scene videos. 
Notably, experiments in BEV segmentation and 3D object detection demonstrate that our synthesized videos provide a substantial performance enhancement to the state-of-the-art BEV perception models, thereby validating its practical applications. 

\end{itemize}

%% file: sec/2_related_work.tex
\section{Related work}
\label{sec:related_work}

\subsection{Video Generation}
Video generation has consistently attracted attention in the field of computer vision. 
Early studies primarily relied on autoregressive models~\cite{wu2021godiva, liang2022nuwa, hong2022cogvideo}.
In recent years, diffusion models~\cite{ho2020denoising, song2020score} have emerged as a powerful alternative in generative tasks. 
Building on the success of diffusion models in image synthesis, multiple attempts have been made to develop large-scale video diffusion models.
These diffusion-based video generation methods~\cite{wang2023modelscope, blattmann2023align, wang2023lavie, chen2024videocrafter2, chen2023videocrafter1, wang2023videofactory, zhang2023controlvideo} typically incorporate temporal interaction modules into image diffusion models to handle temporal information.
The Diffusion Transformer (DiT)~\cite{peebles2023scalable} combines the denoising mechanism of diffusion models with the long-range dependency modeling capabilities of Transformers, demonstrating significant advantages in single-view video generation tasks~\cite{yang2024cogvideox, sora}. 
While DiT has proven effective and extensible in image and single-view video generation, its application to multi-view scenarios remains relatively unexplored. 
In this work, we extend DiT to multi-view video generation, aiming to leverage its strengths in modeling complex, dynamic scenes from multiple viewpoints.

\subsection{Multi-view Scene Generation}

Early methods~\cite{swerdlow2024street} are based on autoregressive models and introduce attention mechanisms incorporating spatial embeddings and camera poses, ensuring that the generated images maintain consistency across different viewpoints.
With the success of diffusion models in generative tasks, researchers begin applying them to multi-view scene generation tasks.
BEVControl~\cite{yang2023bevcontrol} employs the classic UNet structure, inputting both street scene maps and noisy images into the diffusion model, generating multi-view street scenes through iterative denoising.
Building upon these advancements, methods have extended their focus to generate multi-view videos to capture temporal dynamics alongside spatial consistency. 
DriveDreamer~\cite{wang2023drivedreamer}, for example, utilized a multi-stage training process to generate driving videos, aiming to model the complex temporal relationships inherent in dynamic scenes. 
Panacea~\cite{wen2024panacea} and DrivingDiffusion~\cite{li2023drivingdiffusion} integrate intra-view, cross-view, and cross-frame attention modules into the multi-view and temporal models, respectively, and achieve multi-view video generation through a two-stage inference process. 
MagicDrive-V~\cite{gao2024magicdrive3d} introduces decoupled 4D attention into a single network via a two-step training process, enabling multi-view video generation in a single stage.

These decoupled-based methods may face significant limitations in addressing the coupled challenges of cross-frame and cross-view consistency.
To tackle these issues, we propose a novel diffusion transformer network with holistic-4D attention to simultaneously model spatial, temporal, and view dimensions. 

%% file: sec/3_method.tex
\section{Preliminary}
\label{sec:preliminaries}

\begin{figure*}[h]
\centering
\includegraphics[width=\textwidth]{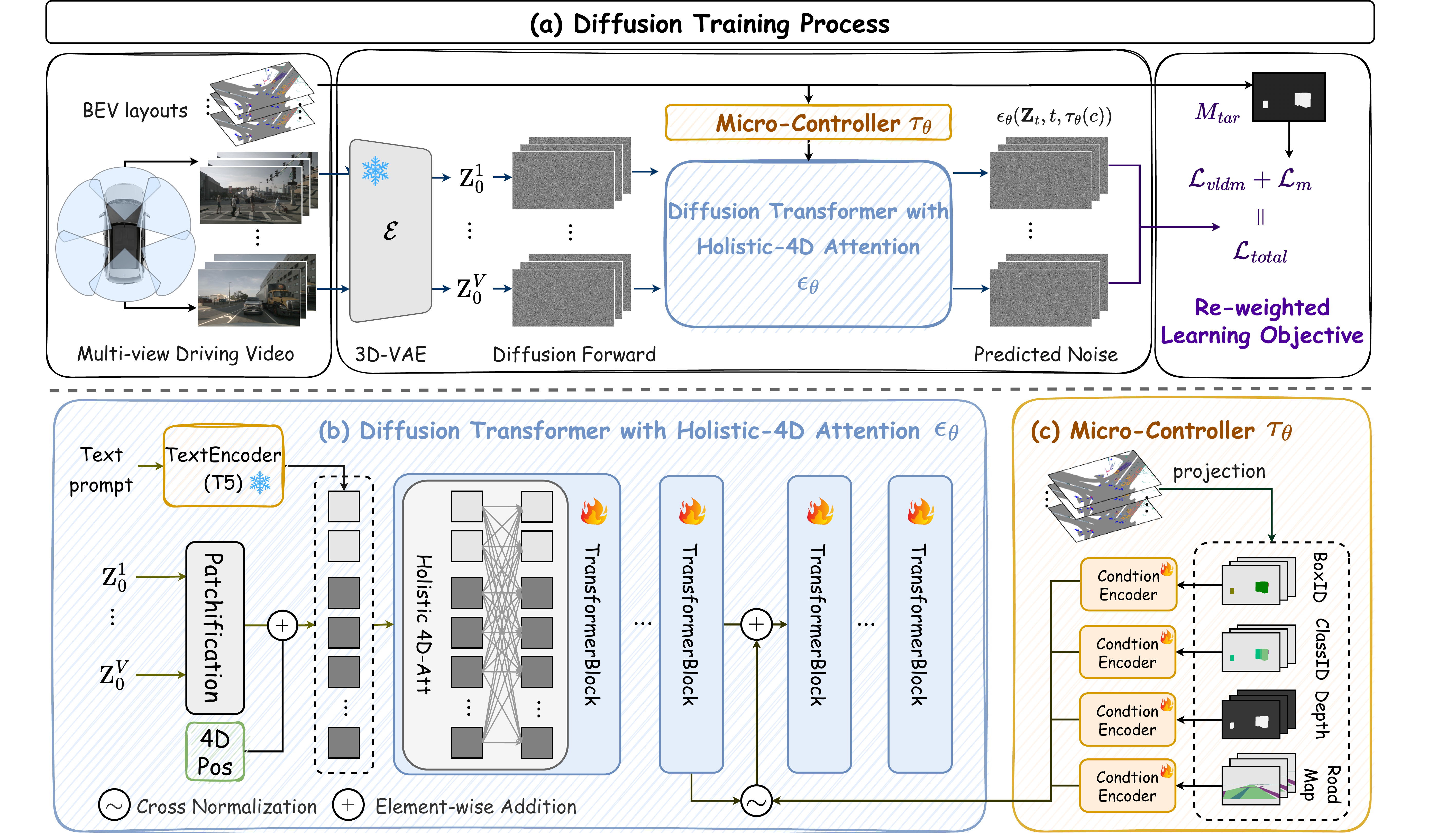}
\vspace{-4mm}
\caption{Overview of our CogDriving. (a) depicts the training process of CogDriving, facilitated by the diffusion transformer with Holistic-4D Attention under the condition of BEV layouts. (b) illustrates the detailed architecture of the diffusion transformer, especially the Holistic-4D Attention to achieve the spatial-temporal-perspective mutual interaction. (c) shows the proposed Micro-Controller for the integration of various conditions.}
\label{fig:overview}
\end{figure*}

\subsection{Problem Formulation}
This work is dedicated to addressing the task of controllable multi-view video generation for autonomous driving. 
Given an ego-vehicle, BEV layouts describe the scene of a video sequence with $T$ frames as $\mathcal{S} =\{{\mathbf{M}}_t, {\mathbf{B}}_t, {\mathbf{P}}_t, {\mathbf{L}}\}_{t=1}^{T}$~\cite{gao2023magicdrive, gao2024magicdrive3d, li2023drivingdiffusion}.
Here, ${\mathbf{M}}_t \in \{0,1\}^{w \times h \times c}$ denotes the binary BEV road map, representing a $w\times h$ meter road layout with $c \in \mathcal{C}$ semantic classes.
${\mathbf{B}}_t = \{ ({\mathbf{b}}_i, c_i) \}_{i=1}^{N}$ represents the 3D bounding boxes for $N$ objects in the scene, where ${\mathbf{b}}_i = \{(x_j, y_j, z_j)\}_{j=1}^{8} \in \mathbb{R}^{8 \times 3}$ denotes the 3D position and $c_i$ is the corresponding class for each object.
${\mathbf{P}}_t=\{{\mathbf{K}}_{v} \in \mathbb{R}^{3 \times 3}, {\mathbf{R}}_{v} \in \mathbb{R}^{3 \times 3}, {\mathbf{T}}_{v} \in \mathbb{R}^{3 \times 1}\}_{v=1}^{V}$ specifies the camera pose, including intrinsics, rotation and translation respectively, for performing transformations from different views, and $\mathbf{L}$ is the scene-level textual description that provides contextual information.
Then, the goal is to generate multi-view driving videos $\mathcal{O} = \{ {\mathbf{O}}_v\}_{v=1}^{V} \in {\mathbb{R}}^{V \times T \times H \times W \times 3} $ that conform to the specified BEV layout $\mathcal{S}$.

\subsection{Video Latent Diffusion Models}
Diffusion Models (DM) generate data by iteratively denoising samples from a noise distribution. 
To simplify the challenge of handling high-dimensional data in pixel space, Latent diffusion models (LDM) are introduced as a variant to operate the diffusion process in the latent space. 
To apply the LDM paradigm to video generation, VLDM~\cite{blattmann2023align} further presents video latent diffusion models by fine-tuning the pre-trained image DM with additional temporal layers. 
Specifically, it involves a variational autoencoder (VAE) that includes an encoder $\mathcal{E}$ and a decoder $\mathcal{D}$, a denoising network $\epsilon_{\theta}$, and a condition encoder $\tau_{\theta}$. 
Given a video sample $\mathbf{x}_0 \in \mathbb{R}^{T \times H \times W \times 3}$ containing $T$ frames, VLDM first encodes it to get a low-dimensional latent representation $\mathbf{Z}_0 \in \mathbb{R}^{T \times H' \times W' \times C}$ using $\mathcal{E}(\cdot)$. 
Then, after achieving the optional condition encoding $\tau_{\theta}(c)$, the denoising process $\epsilon_{\theta}$ can be learned by the following objective:
\begin{equation}
\label{eq:vldm}
    \mathcal{L}_{vldm} = \mathbb{E}_{\mathcal{E}(\mathbf{x}),\epsilon,t,\tau_{\theta}(c)}
\left[ \|\epsilon - \epsilon_{\theta}(\mathbf{Z}_t,t,\tau_{\theta}(c))\|^{2}_{2} \right],
\end{equation}
where $t$ is the time-step, $\mathbf{Z}_t = \alpha_t \mathbf{Z}_0 + \beta_t \epsilon$ is the diffused inputs in the latent space at time step $t$, $\epsilon \in \mathcal{N}(\mathbf{0}, \mathbf{I}) $ is additive Gaussian noise, and $c$ refers to the conditional information provided to the model, which could be a text prompt, an image, or any other user-specified condition. 
Once trained, VLDM is able to generate new data by iterative denoising a random Gaussian noise in the latent space. The denoised latents are then converted back to a video by $\widetilde{\mathbf{x}}=\mathcal{D}(\cdot)$.

\section{Method}
\label{sec:method}
\subsection{Overview}
In this section, we introduce the CogDriving in detail, and present the overview in Fig.~\ref{fig:overview}. 
The CogDriving proposes a simple yet effective DiT-based network, equipped with holistic-4D attention to simultaneously model spatial, temporal, and view dimensions.
For a compact latent space, CogDriving explores 3D-VAE in the network to encode multi-view videos.
To integrate various types of conditioning controls, we design the Micro-Controller for scene-level controllability inspired from ControlNeXt~\cite{peng2024controlnext}.
The Micro-Controller consists of multiple independent lightweight condition encoding modules and Cross Normalizations, detailed in Sec.~\ref{sec:4.3}.
Finally, a re-weighted learning objective to rectify the learning imbalance and enhance the capability and quality of instance generation is presented in Sec~\ref{subsec:loss}.

\subsection{DiT with Holistic-4D Attention}
\label{sec:4.1}
\noindent{\textbf{3D-VAE for Multi-view Encoding.}}

Inspired by the success of Sora~\cite{sora} and CogVideoX~\cite{yang2024cogvideox} on video generation, we extend the 3D Variational Autoencoders (3D-VAE) to multi-view video generation to achieve a compact video representation. 
Single-view video generation models~\cite{sora, yang2024cogvideox} apply 3D-VAE to compress both the spatial and temporal dimensions simultaneously.
In multi-view video data, the redundancy between different camera views is relatively low. 
Even between adjacent cameras, overlapping regions are confined to the peripheries of their respective fields of view.  
Therefore, we opt not to compress the view dimension. 
Instead, we independently encode the data from each camera view using a shared 3D-VAE, allowing us to preserve the unique information from each view while benefiting from the compact representation.
In practice, for a multi-view video sample $\mathbf{x} \in \mathbb{R}^{V\times T\times H\times W\times 3}$, each view $\mathbf{x}_v\in \mathbb{R}^{T\times H\times W\times 3}$ is mapped to latent feature individually via 3D-VAE encoder:
\begin{equation}
    \mathbf{Z}_v=\mathcal{E}(\mathbf{x}_v), v=1,...,V.
\end{equation}

\noindent{\textbf{Patchification.}} 
Before feeding the video latents into the DiT for denoising, patchifying is essential to convert these latent representations into a sequence of tokens. 
Given the latent feature of each view $\mathbf{Z}_v\in \mathbb{R}^{T'\times H'\times W'\times C}$, we implement patchification operation with a patch size of $p \times p$ to create a sequence of patch tokens $\mathbf{p}_i\in \mathbb{R}^{L\times C}$, where $L\!=\!\frac{T'\times H'\times W'}{p^2}$ represents the total number of patchified tokens. 
A 2D convolution layer is subsequently applied to these patch tokens to derive patch embeddings $\mathbf{E}_{v} \in \mathbb{R}^{L\times D}$, with $D$ denoting the embedding dimension.

\noindent{\textbf{Positional Encoding.}} 
Compared to single-view video generation, the patch embeddings of multi-view video data need to capture additional positional information for the view dimension in order to ensure coherence across different views.
To accommodate the pre-trained weights from~\cite{yang2024cogvideox}, we treat the view dimension as an extension of the temporal dimension, enabling the use of the original 3D positional encoding scheme from the video generation training process.
Referring to the frequency-based sine-cosine 3D positional encoding function in VisTR~\cite{wang2021end}, \ie, $PE_{3D}$, we define our 4D positional encoding as:
\begin{equation}
    \mathrm{PE_{4D}}(v, t, h, w) = \mathrm{PE}_{3D}(t + (v - 1) * T', h, w),
\end{equation}
where $v, t, h, w$ denote the indexes of view, temporal, and spatial dimensions, respectively.

\noindent{\textbf{Holistic-4D Attention.}} 
When the DiT is employed to process video data, previous methods~\cite{sora, yang2024cogvideox} have considered only spatial and temporal attention mechanisms. 
To adapt DiT for multi-view videos, achieving cross-view consistency is essential, necessitating the incorporation of the view dimension into the transformer architecture.
Several approaches have attempted to address this by adding additional attention layers to handle the view dimension separately. 
However, as previously discussed, this method requires extensive implicit transmission of visual information across multiple attention layers, increasing complexity and making it more challenging to achieve coherent and consistent results.
Here, we propose a simple yet effective holistic-4D attention mechanism. 
Specifically, we concatenate the patch embeddings from all views along the sequence dimension to form a unified input for the transformer.
Following the design philosophy of the base network~\cite{yang2024cogvideox} to better align visual and semantic information, we prepend the text embeddings $\mathbf{E}_{txt}\in \mathbb{R}^{L_{txt}\times D}$ to the sequence of patch embeddings, where the length of the text embedding $L_{txt}$ is decided by the text encoder~\cite{2020t5}.
Thereby, we obtain the input embeddings for the attention module $\mathbf{E}=[\mathbf{E}_{txt}; \mathbf{E}_{1}; \ldots; \mathbf{E}_{V}] \in \mathbb{R}^{(L_{txt} + L\cdot V)\times D}$.
We then perform holistic-4D self-attention across all frames and views, where each embedding $\mathbf{e}_i = \mathbf{E}_{i:} \in \mathbb{R}^{1 \times D}$ can attend to all embeddings as,
\begin{equation}
    \mathrm{Att}_{4D}(Q_i,K,V) = \!\!\!\! \sum_{j=1}^{L\cdot V + L_{txt}} \!\!\!\! \mathrm{Softmax}(\frac{Q_i\cdot K_j^T}{\sqrt{D_h}})V_j,
\end{equation}
where $Q_i = \mathbf{e}_i W^{Q}, K=\mathbf{E}W^{K}, V=\mathbf{E}W^{V}$ is obtained by linear projections of $\mathbf{E}$ and its $i$-th row vector.

\subsection{Micro-Controller for Condition Controls}
\label{sec:4.3}

Previous approaches typically employ ControlNet to achieve controllable generation by duplicating the backbone network to process conditional inputs. 
However, in our case, experiments reveal that duplicating the 4D attention branch leads to significant challenges in convergence and training stability, largely due to the substantial increase in parameters and computational overheads. 
To enable controllability, we are inspired by ControlNeXt~\cite{peng2024controlnext} and propose a lightweight design of the condition branch, \ie, a micro-controller, tailored to multi-view generation tasks.

To make conditions more manageable for the micro-controller, we first preprocess the 3D descriptions $\mathcal{S}$ of the video sequence. 
For each frame, we project the binary BEV road map $\mathbf{M}_t$ and 3D bounding box $\mathbf{B}_t$ onto each camera view according to the camera pose $\mathbf{P}_t$, resulting in \underline{\textit{road map}}, \underline{\textit{class ID}}, \underline{\textit{box ID}}, and \underline{\textit{depth map}} as the conditional inputs, each of which has the size of $T\times V \times H \times W \times 1$.
The controller then employs four separate encoders to process each condition individually, generating corresponding condition embeddings, \ie, $\mathbf{E}_{R},\mathbf{E}_{C}, \mathbf{E}_{B}$ and $\mathbf{E}_{D}$. 
Each encoder shares a uniform architecture consisting of alternating convolutional and pooling layers, aligning the conditional information with the spatial and temporal dimensions of the video data in latent space.

We integrate the conditional feature with the video latent feature at the intermediate layer of the network. 
To ensure compatibility, we apply a cross-normalization technique~\cite{peng2024controlnext} to each conditional embedding respectively to align their distributions with the video features. 

\begin{figure}
    \centering
    \includegraphics[width=0.7\linewidth]{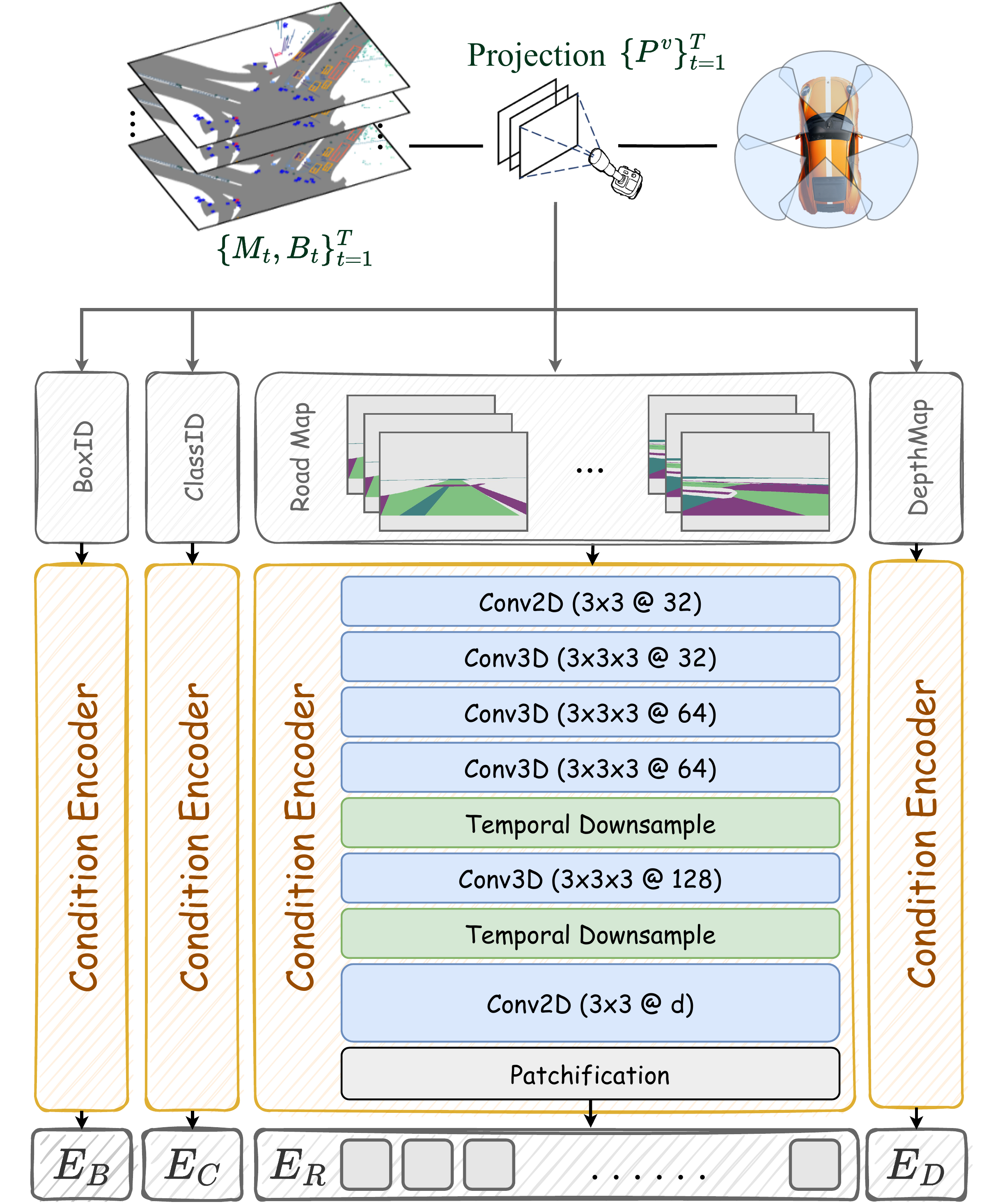}
    \vspace{-1mm}
    \caption{Our lightweight Micro-Controller encodes road maps, box IDs, class IDs, and depth maps independently from 3D annotations for precise, geometry-guided synthesis.}
    \label{fig:micro_controller}
\end{figure}

\subsection{{Re-weighted Learning Objective}}
\label{subsec:loss}
The learning objective of the VLDMs typically assigns equal importance to all pixels in the video, ensuring uniform learning across frames. 
However, in the context of driving street scenes, key elements for scene understanding and decision-making are typically small objects, such as traffic cones, pedestrians, distant vehicles $etc.$ 
Therefore, their importance necessitates high-quality generation of these small objects.
In practice, training diffusion models with equal weights often results in insufficient learning for these objects, as they may be overshadowed by larger and more prominent contents.
To address this issue, we propose a re-weighted learning objective.

Specifically, we utilize the encoded target category information in Sec~\ref{sec:4.3} to extract a target mask $M_{tar}\in \mathbb{R}^{V\times T\times H\times W\times 1}$ as:
\begin{equation}
    M_{tar}(v, t, h, w)=\left\{\begin{matrix}
                         1 & R(v, t, h, w) > 0 \\
                         0 & R(v, t, h, w) = 0
                        \end{matrix}\right.
\end{equation}
which identifies the regions containing objects of interest.
With this target mask, we re-weight the error between the predicted and target noises as follows:
\begin{equation}
\begin{aligned}
    \mathcal{L}_{m} \!=\! \mathbb{E}_{\mathcal{E}(x),\epsilon,t,\tau_{\theta}(c)} \! \left[\left\|\frac{M_{tar} \! \cdot \! (\epsilon \!-\! \epsilon_{\theta}(\mathbf{Z}_t,t,\tau_{\theta}(c)))}{\sum{M_{tar}}}\right\|_2^2\right]
\end{aligned}
\end{equation}
and then combine it with the VLDM objective in Eq.(\ref{eq:vldm}), achieving the total loss function $\mathcal{L}_{total} = \mathcal{L}_{vldm} + \lambda\cdot\mathcal{L}_m$, where we set $\lambda$ to 0.1 in practice.
This re-weighting strategy encourages the model to focus more on accurately generating small and essential objects during training. 

\begin{table*}[htbp]
\setlength{\abovecaptionskip}{3pt} 
\setlength{\belowcaptionskip}{0pt}
\setlength{\tabcolsep}{1.5mm}
\centering
\small
\caption{Comparison with state-of-the-art methods on nuScenes validation set in \textit{generation quality} and \textit{controllability}. $\uparrow / \downarrow$ indicates that a higher/lower value is better. 
The 'Oracle' presents the test results of corresponding models trained on nuScenes training set.
}
\label{tab:cmp_sota}
\renewcommand{\arraystretch}{1.1}
\resizebox{0.95\textwidth}{!}{
    \begin{tabular}{@{}l|cccc|cc|cc@{}}
        \toprule
         \ & \multicolumn{4}{c|}{Generation quality} & \multicolumn{4}{c}{Controllability} \\
         \ & \multicolumn{4}{c|}{} & \multicolumn{2}{c|}{BEV segmentation} & \multicolumn{2}{c}{3D object detection} \\     
         \cline{2-9}
         Method & M-View & M-Frame  & FID ($\downarrow$) & FVD ($\downarrow$) & Road mIoU ($\uparrow$) & Vehicle mIoU ($\uparrow$) & Drivable mIoU ($\uparrow$) & Object NDS($\uparrow$) \\
        \midrule
         Oracle        & \cellcolor{gray!30}     & \cellcolor{gray!30}      & \cellcolor{gray!30}    & \cellcolor{gray!30}    & \cellcolor{dcolor!40}{71.6}   & \cellcolor{dcolor!40}{35.8}  & \cellcolor{dcolor!40}{81.7}  & \cellcolor{dcolor!40}{41.2}   \\        
        BEVGen~\cite{swerdlow2024street} & \checkmark     &      & 25.5 &-   & 50.2 (-21.4\%)      & 5.9 (-29.9\%)          &-   &-    \\
        BEVControl~\cite{yang2023bevcontrol}             & \checkmark     &             & 24.9 &-   & 60.8 (-10.8\%)      & 26.8 (-9.0\%)         &-   &-    \\
        MagicDrive~\cite{gao2023magicdrive}             & \checkmark     &             & 16.2 &-   & 61.1 (-10.5\%)      & 27.0 (-8.8\%)         &-              & 30.6 (-10.6\%)    \\
        DriveDreamer~\cite{wang2023drivedreamer}           &            & \checkmark           & 52.6 & 452.0   &-    &-   &-   &-    \\
        \midrule
        Panacea~\cite{wen2024panacea}                & \checkmark          & \checkmark           & 17.0 & 139.0    &-    &-   &-   &-    \\
        DrivingDiffusion~\cite{li2023drivingdiffusion}       & \checkmark     & \checkmark    & 15.8  & 346.0     & 63.2 (-8.4\%)  & 31.6 (-4.2\%)      & 67.8 (-13.9\%)      & 33.1 (-8.1\%)  \\
        MagicDrive-V~\cite{gao2024magicdrive3d} & \checkmark          & \checkmark           & 20.7 & 164.7    & 40.0 (-31.6\%)    & 22.9 (-12.9\%)   &38.6 (-43.1\%)   &19.6 (-21.6\%)    \\
        \rowcolor{dcolor} CogDriving (Ours)      & \checkmark          & \checkmark           & 15.3 & 37.8     & 65.7 (-3.7\%)     & 32.1 (-3.7\%)        & 71.9 (-9.8\%)         & 34.3 (-6.9\%)  \\
        \bottomrule
    \end{tabular}
}
\end{table*}

%% file: sec/4_experiment.tex
\section{Experiment}

\subsection{Experiment Details}

\noindent{\textbf{Dataset.}}
To verify the performance of CogDriving in multi-view video generation for autonomous driving, we conduct experiments on the public {nuScenes}~\cite{caesar2020nuscenes} dataset, a prevalent dataset in BEV segmentation and detection for driving.
The nuScenes dataset contains 1000 video sequences, each with a length of approximately 20 seconds in length, captured from six cameras.
Following the official configuration, we utilize 700 street-view videos for training, 150 for validation and 150 for testing. 

\noindent{\textbf{Training.}}
To facilitate training, we segment and resize the training split of nuScenes to generate video clips with a duration of 16 frames and a spatial resolution of $720 \times 480$.
We initialize our DiT with 4D attention using pre-trained CogVideoX~\cite{yang2024cogvideox} and train it on the NuScenes dataset.
The entire training process is divided into two phases using eight NVIDIA A100 GPUs. 
In the first phase, we fine-tune all parameters on the nuScenes dataset for $10^{4}$ iterations with a learning rate of $1\times 10^{-5}$.
In the second phase, we continue fine-tuning using LoRA~\cite{hu2021lora} while jointly training the micro-controller module, for an additional $1.5\times 10^{4}$ iterations with a learning rate of $5\times 10^{-4}$.
Once trained, our final model generates multi-view videos directly according to the provided conditions in a single stage, without requiring post-processing.

\noindent{\textbf{Evaluation Metrics.}}
The performance evaluation of multi-view video generation is conducted from two key aspects: generation quality and controllability.
To measure generation quality, we use the frame-wise Fréchet Inception Distance (FID)~\cite{FID} and Fréchet Video Distance (FVD)~\cite{FVD} metrics.
For controllability, we compare the validation performance of our generated data against real data through two perception tasks: BEV segmentation and 3D object detection, using CVT~\cite{zhou2022cross} and BEVFusion~\cite{liu2023bevfusion} as the perception models respectively. 

\subsection{Quantitative Analysis}
In order to ensure a fair and comprehensive comparison, we generate the complete validation set of the nuScenes dataset and compute quantitative metrics to evaluate CogDriving against existing state-of-the-art approaches, including multi-view image-based methods, \eg, BEVGen~\cite{swerdlow2024street}, BEVControl~\cite{yang2023bevcontrol},  MagicDrive~\cite{gao2023magicdrive}, single-view video-based methods, \eg, DriveDreamer~\cite{wang2023drivedreamer}, and multi-view video-based methods, \eg, MagicDrive-V~\cite{gao2024magicdrive3d}, DrivingDiffusion~\cite{li2023drivingdiffusion}, Panacea~\cite{wen2024panacea}.

\noindent{\textbf{Generation Quality.}}
As shown in Table~\ref{tab:cmp_sota}, our CogDriving outperforms other existing state-of-the-art methods in both image and video qualities, yielding notably lower FID and FVD scores. 
Since the FID metric measures image quality, the improvement in FID is relatively modest compared to other methods.
However, it is noteworthy that CogDriving showcases significantly superior generation quality with an FVD of 37.8 compared with all video-based generation methods, indicating exceptional temporal coherence and dynamic realism, and setting a new benchmark for multi-view video generation quality.

\noindent{\textbf{Controllability.}}
Following~\cite{gao2023magicdrive, li2023drivingdiffusion}, we evaluate the controllability of CogDriving using two pre-trained perception models: CVT~\cite{zhou2022cross} and BEVFusion~\cite{liu2023bevfusion} for BEV segmentation and 3D object detection tasks, and report the results in Table~\ref{tab:cmp_sota}.
For comparison, we apply the pre-trained CVT and BEVFusion on real data directly to obtain the Oracle performance, which serve as indicators of how well the generated samples align with the conditioned BEV layout sequences.
Subsequently, we employ the same models to process the validation set generated by various generative models.
As shown in Table~\ref{tab:cmp_sota}, CogDriving demonstrates excellent performance in BEV segmentation, with a degradation of only 3.7\% in both Road and Vehicle mIoU metrics compared to the Oracle, indicating effective alignment with the provided road map conditions.
For 3D object detection, CogDriving also attains favorable NDS values, indicating robust controllability in generating object instances that accurately reflect the conditioned bounding box information.

\noindent{\textbf{Synthetic Data for Training.}}
Beyond the validation set evaluation, we further investigate the data augmentation capabilities of CogDriving by synthesizing data to support the training of BEV segmentation and 3D object detection models.
We generate multi-view images in quantities matching those in the original nuScenes training set and train CVT~\cite{zhou2022cross} and BEVFusion~\cite{liu2023bevfusion} with this augmented dataset.
As shown in Tab.~\ref{tab:support_train}, incorporating synthetic data generated by CogDriving into the training process enhances the segmentation performance of CVT, resulting in improved accuracy for both drivable areas and dynamic vehicle classes. 
For 3D object detection, as presented in Tab.~\ref{tab:support_train}, the model trained on the augmented dataset achieves higher NDS and mAP values compared to the model trained solely on real data.
These improvements confirm the framework's capability to produce high-fidelity multi-view video samples, which effectively augments training data and enhances the performance of autonomous driving systems.
\subsection{Qualitative Analysis}
\label{subsec:qualitative}
\begin{figure}[t]
    \centering
    \includegraphics[width=0.98\linewidth]{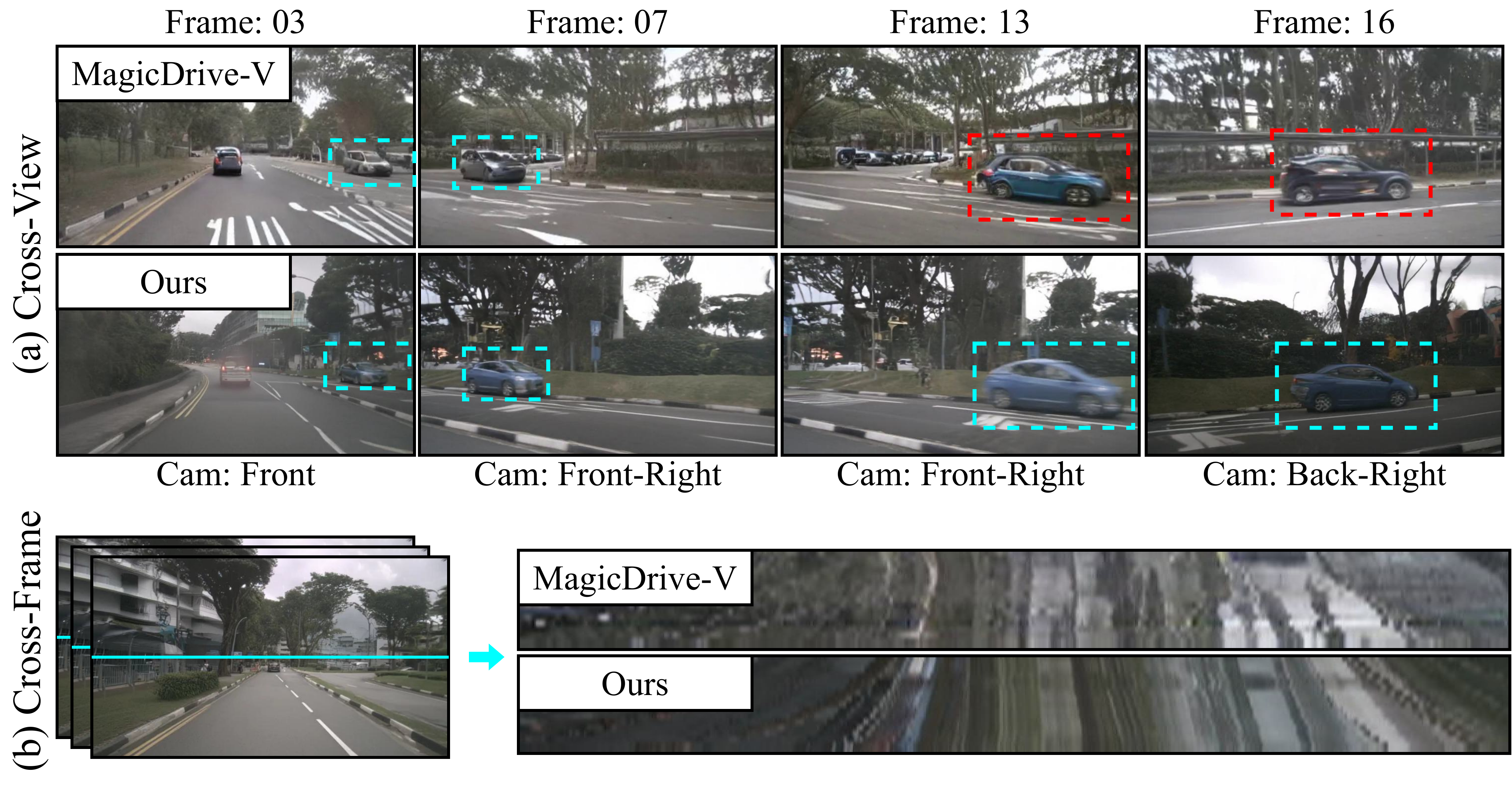}
    \vspace{-4mm}
    \caption{Consistency analysis. 
    When the same object appears at different times and views, CogDriving maintains cross-view consistency.
    According to the temporal profile, CogDriving shows superior cross-frame consistency with continuous lines.}
    \label{fig:cmp_sota}
\end{figure}

\noindent{\textbf{Consistency.}}
We demonstrate the consistency of the proposed CogDriving in Fig~\ref{fig:cmp_sota}.
MagicDrive-V~\cite{gao2024magicdrive3d}, as a single-stage video generation model, is the most closely related to our approach and is the only method that has publicly released complete model weights. 
We utilize the official weights of MagicDrive-V to generate videos and conduct a comparative analysis of the generated quality.
The Fig~\ref{fig:cmp_sota}(a) illustrates a fast-moving car appearing in four different viewpoints.
Previous method~\cite{gao2024magicdrive3d} based on decoupled attention struggles to establish consistent associations when handling objects appearing at different times across views, resulting in noticeable changes in the appearance of the car. 
In contrast, our framework with holistic-4D attention successfully captures dependencies across multiple dimensions, maintaining cross-view consistency of the object.

To further validate the cross-frame consistency of our CogDriving, we follow the approach used in FRVSR~\cite{sajjadi2018frame} to create a temporal profile, as shown in Fig~\ref{fig:cmp_sota}(b).
Any temporal flickering in the video will manifest as jitter or jagged lines in the temporal profile. 
As seen in Fig.\ref{fig:cmp_sota}, the temporal profile produced by CogDriving exhibits more smooth and continuous lines, demonstrating the excellent cross-frame consistency of the generated video.

\begin{table}[t]
\setlength{\abovecaptionskip}{3pt} 
\setlength{\belowcaptionskip}{0pt}
\setlength{\tabcolsep}{1.5mm}
\centering
\small
\caption{
The BEV segmentation and detection models are trained on multi-view data synthesized by CogDriving, and evaluated on the NuScenes validation split.
}
\label{tab:support_train}
\renewcommand{\arraystretch}{1.1}
\resizebox{0.48\textwidth}{!}{
\hspace{-4mm}
\begin{tabular}{l|cc|cc}
\toprule
\multirow{2}{*}{Data} & \multicolumn{2}{c|}{BEV Segmentation} & \multicolumn{2}{c}{3D Object Detection} \\ \cline{2-5} 
                      & Road mIoU ($\uparrow$)       & Vehicle mIoU ($\uparrow$)       & mAP ($\uparrow$)                & Object NDS ($\uparrow$)                \\
\midrule
w/o synthetic data    & 71.6            & 35.8               & 32.9               & 37.8               \\
w/ MagicDrive         & 77.5            & 38.4               & 34.3               & 38.2               \\
\rowcolor{dcolor} w/ CogDriving         & 79.9            & 40.4               & 35.7               & 39.7              \\
\bottomrule
\end{tabular}
}
\end{table}

\begin{figure*}[t]
    \centering
    \includegraphics[width=0.96\linewidth]{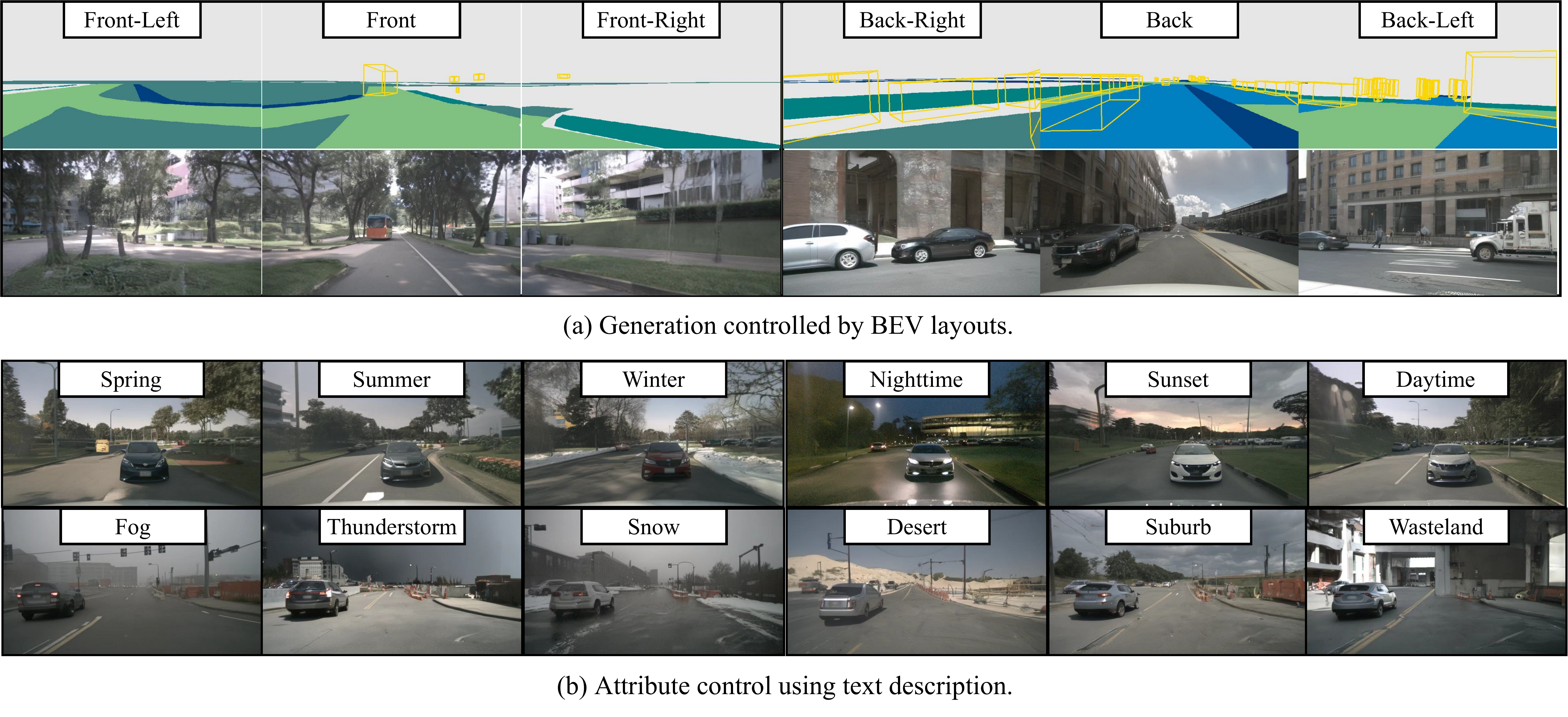}
    \vspace{-4mm}
    \caption{Generation results of CogDriving.
    (a). CogDriving synthesizes multi-view driving scene videos conditioned on Bird-Eye-View (BEV) layout sequences. 
    (b). The model demonstrates its strong generalization capability by generating diverse driving videos, including different weather, seasons, times, and extreme scenarios such as thunderstorms.}
    \label{fig1}
\end{figure*}

\noindent{\textbf{Attribute and Layout Control.}}
CogDriving can generate consistent multi-view content based on control inputs.
In Fig.~\ref{fig1}(a), our approach accurately generates road structures (curved lanes in Front-Left view) and objects (pedestrians, vehicles) based on the provided BEV layouts.
In Fig.~\ref{fig1}(b), we demonstrate how modifications to textual prompts allow us to manipulate various scene attributes, including weather conditions, season, time of day, and specific scene characteristics. 
This includes the ability to simulate extreme weather events, such as thunderstorms, as well as rare scenarios like desert landscapes.
This flexibility significantly enhances the diversity of generated data, facilitating the creation of varied and complex driving environments.
More results are provided in the supplementary materials.

\begin{table}
\setlength{\abovecaptionskip}{3pt} 
\setlength{\belowcaptionskip}{0pt}
\setlength{\tabcolsep}{1.5mm}
\centering
\small
\caption{Ablation of the holistic 4D-attention module, the micro-controller, and the re-weighted learning objective.
Columns painted with blue color represent the final model configuration.
}
\label{tab:ablation}
\renewcommand{\arraystretch}{1.2}
\resizebox{0.38\textwidth}{!}{
\begin{tabular}{l|ccc}
\toprule
                    & FID($\downarrow$)  & FVD($\downarrow$)  & Object NDS($\uparrow$) \\
\midrule
\rowcolor{dcolor} Holistic-4D         & 15.3 & 37.8 & 34.3        \\
Decouple ST-V    & 15.8 & 69.1 & 32.3        \\
\midrule
Cond\_ControlNet          & 24.9  & 80.3  & 29.1  \\
Micro-control-$5$     & 156.1  & 563.2  & 0.2 \\
\rowcolor{dcolor} Micro-control-$10$     & 15.3  & 37.8  & 34.3  \\
Micro-control-$15$     & 15.9  & 44.2  & 22.6  \\
\midrule
w/o $\mathcal{L}_{m}$         & 15.6  & 40.0  & 30.6  \\
\rowcolor{dcolor} w $\mathcal{L}_{m}$      & 15.3  & 37.8  & 34.3  \\
\bottomrule
\end{tabular}
}
\end{table}

\subsection{Ablation Studies}
\label{sebsec:ablation}
\begin{figure}[]
    \centering
    \includegraphics[width=0.96\linewidth]{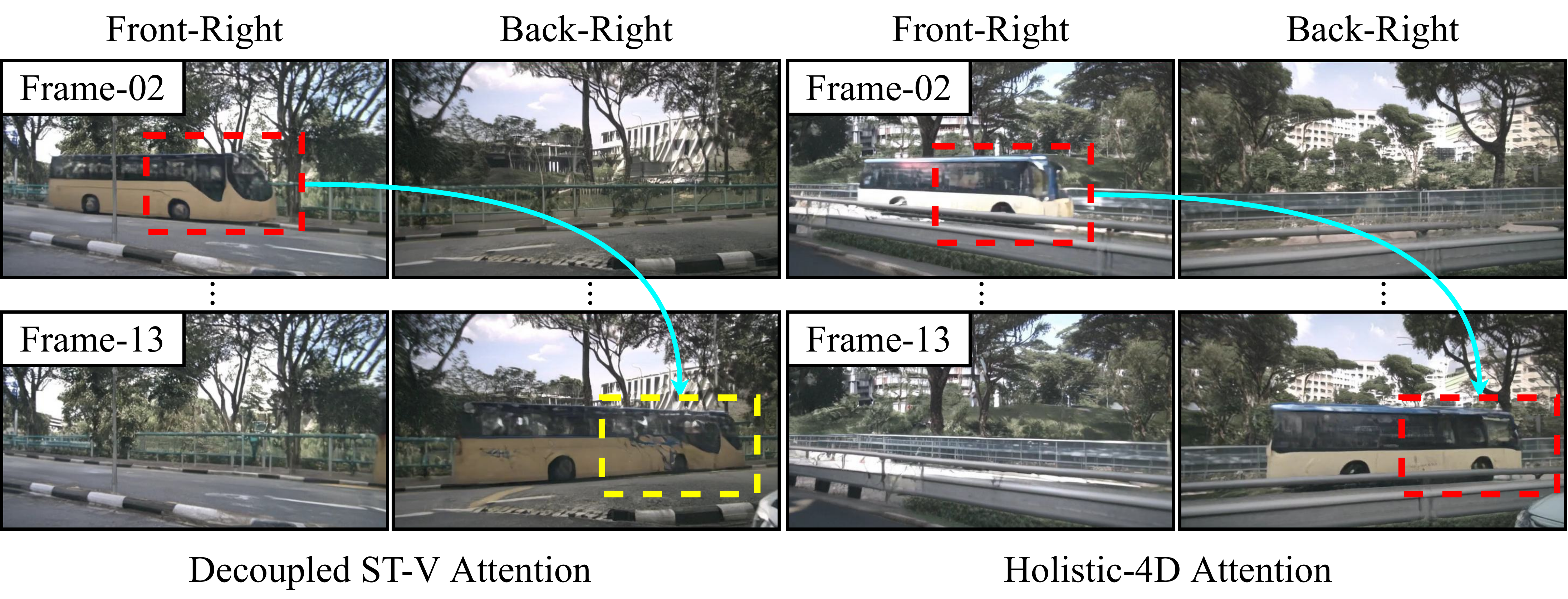}
    \vspace{-3mm}
    \caption{Ablation on the efficiency of holistic-4D attention.
    The {\color{cyan1}cyan lines} connect the same part of the object across different frames and views for consistency comparison.
    }
    \label{fig:ablation_4d_att}
\end{figure}

\noindent{\textbf{Holistic-4D Attention.}}
To evaluate the impact of holistic-4D attention, we train a comparative model where the view dimension is decoupled from the spatial and temporal attention mechanisms (the Decouple ST-V model).
As shown in Tab.~\ref{tab:ablation}, the model utilizing holistic 4D attention achieves a higher FVD score, indicating improved consistency, which in turn contributes to an increase in the NDS metric. 
In Fig.~\ref{fig:ablation_4d_att}, the network based on holistic 4D attention maintains strong cross-view consistency. 
In contrast, the decoupled view attention model exhibits noticeable appearance discrepancies in the generated objects across different views.
Compared to the decoupled attention, although our holistic-4D attention mechanism is slightly slower, it remains competitive in terms of efficiency as reported in Table~\ref{tab:compute_time}. Moreover, benefiting from the holistic-4D attention, our approach can generate higher quality multi-view videos and improve the results significantly.

\begin{figure}[]
    \centering
    \includegraphics[width=0.96\linewidth]{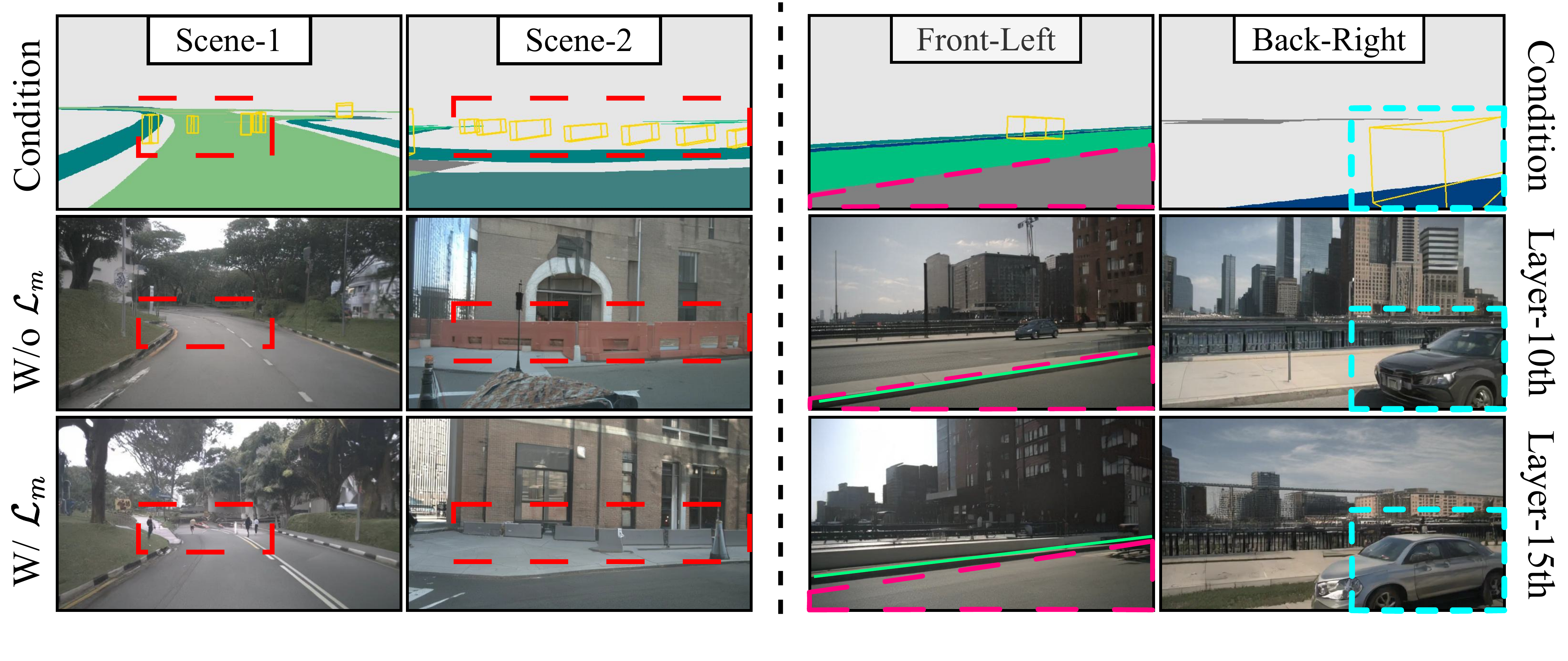}
    \vspace{-4mm}
    \caption{
    Ablation on the position of conditional injections of our Micro-Controller, and the re-weighted learning objective.
    }
    \label{fig:ablation_pos_loss}
\end{figure}

\noindent{\textbf{Micro-Controller.}}
The Micro-Controller is a lightweight controller designed to enable scene-level controllability tailored to our holistic-4D attention-based backbone.
In Tab.~\ref{tab:ablation}, we compare a variant of CogDriving, in which its Micro-Controller is replaced by the traditional ControlNet, referred to as Cond\_ControlNet. 
Notably, the Micro-Controller contains $\sim$10.53M parameters, which is only 1.1\% of the parameter count of ControlNet, however, it demonstrates superior performance in both generation quality and control capability.
Moreover, we explore the optimal position for conditional information injection by analyzing the variants Micro-control-$*$, where $*$ denotes the layer index of DiT.
From Tab.~\ref{tab:ablation} and Fig.~\ref{fig:ablation_pos_loss}, we note that inserting the conditional embeddings at intermediate layers yields the best results. 
Early layer insertion leads to unstable training, while late layer insertion results in insufficient control.
\begin{table}
\setlength{\abovecaptionskip}{3pt} 
\setlength{\belowcaptionskip}{0pt}
\setlength{\tabcolsep}{1.5mm}
\centering
\small
\caption{The computational efficiency comparison between decoupled attention and Holistic-4D attention. The experiments are conducted on an Nvidia A100 GPU.
}
\label{tab:compute_time}
\renewcommand{\arraystretch}{1.2}
\resizebox{0.28\textwidth}{!}{
\begin{tabular}{l|cc}
\toprule
Attention type           & Train      & Inference \\
\midrule
\rowcolor{dcolor} Holistic-4D      & 30.2s/iter & 6.6s/iter \\
Decouple ST-V & 28.9s/iter & 5.7s/iter \\
\bottomrule
\end{tabular}
}
\end{table}

\noindent{\textbf{Re-weighted Learning Objective.}}
From Tab.~\ref{tab:ablation}, one can note that when applying the re-weighted loss, our method achieves a notable improvement in detection performance, reflecting enhanced instance generation quality.
Additionally, as shown in Fig.~\ref{fig:ablation_pos_loss}, the re-weighted learning objective facilitates small object generation in the street scenes, such as pedestrians and obstacles.

%% file: sec/5_conclusion.tex
\section{Conclusion}

In this paper, we introduce a novel network for generating high-quality multi-view videos in autonomous driving systems, \ie, CogDriving. 
To ensure both cross-frame and cross-view consistency, a Diffusion Transformer architecture equipped with holistic-4D attention modules is designed to associate across the spatial, temporal, and viewpoint dimensions.
Moreover, to enable precise control of BEV layouts, we propose a lightweight control module, \ie, Micro-Controller, tailored to CogDriving.
To struggle with the imbalanced distribution between the foreground and background, we incorporate a re-weighted learning objective that dynamically adjusts the importance of object instances during training, and yields the essential object instances generation.
Experimental results demonstrate that CogDriving is capable of generating realistic and consistent videos, which can be utilized as data augmentation for BEV perception algorithms.
We believe that further exploration, incorporating diverse control signals, holds promise for generating more diverse video samples in the future.


%% file: sec/X_suppl.tex
\clearpage
\setcounter{page}{1}

\makeatletter
\def\maketitlesupplementary{
   \begingroup 
       \onecolumn
       \centering
       \Large
       \textbf{\thetitle}\\
       \vspace{0.5em}Supplementary Material \\
       \vspace{1.0em}
   \endgroup
}
\makeatother
\maketitlesupplementary

In Sec~\ref{sec:quant_results}, we provide detailed quantitative results and analysis.
Additionally, we present more comprehensive visualizations and conduct a detailed comparison in Sec~\ref{sec:vis_results}.
Finally, we include specific details regarding the training processes in Sec~\ref{sec:train_detail}.
More video results in mp4 format are included in the supplementary zip file.

\section{Quantitative Analysis: Controllability}
\label{sec:quant_results}
As a supplement to Sec.5.2 and Tab.1 of the main manuscript, we include additional evaluations of the controllability of MagicDrive-V~\cite{gao2024magicdrive3d} and Panacea~\cite{wen2024panacea}, which are state-of-the-art multi-view driving video generation methods.
For MagicDrive-V, we utilize the official pretrained weights to generate videos on the nuScenes validation split.
For Panacea, since the full model weights are not publicly available, we employ the generated samples provided by the authors for our testing.
We assess the controllability of both methods on 3D detection task and BEV segmentation task with BEVFusion~\cite{liu2023bevfusion} models.

The 3D detection results on nuScenes validation set are presented in Tab~\ref{tab:supp_detection}. 
It can be noted that, our method surpasses MagicDrive-V and Panacea in both mean Average Precision (mAP) and the overall metric Object NDS, demonstrating superior object generation capabilities.
The BEV segmentation results are displayed in Tab~\ref{tab:supp_segmentation}.
Our method achieves higher segmentation accuracy across all categories compared to the MagicDrive-V, showcasing excellent controllable generation performance. 

\begin{table*}[h]
\setlength{\abovecaptionskip}{3pt} 
\setlength{\belowcaptionskip}{0pt}
\setlength{\tabcolsep}{2mm}
\centering
\renewcommand{\arraystretch}{1.3}
\small
\begin{tabular}{l|ccccccc}
\toprule
Method    & mAP $\uparrow$            & mATE $\downarrow$           & mASE $\downarrow$           & mAOE $\downarrow$           & mAVE $\downarrow$           & mAAE $\downarrow$           & Object NDS $\uparrow$             \\
\midrule
\rowcolor{gray!30} Oracle~\cite{liu2023bevfusion}       & 0.3554          & 0.6678          & 0.2726          & 0.5612          & 0.8955          & 0.2593          & 0.412           \\
MagicDrive-V~\cite{gao2024magicdrive3d} & 0.0242          & 0.8098          & 0.4167          & 0.5949          & 0.8707          & 0.4738          & 0.1955          \\
Panacea~\cite{wen2024panacea}      & 0.2091          & \textbf{0.6846} & 0.2735          & \textbf{0.5697} & \textbf{0.8644} & \textbf{0.2612} & 0.3392          \\
\rowcolor{dcolor} CogDriving (Ours)         & \textbf{0.2218} & 0.6850           & \textbf{0.2731} & 0.5734          & 0.8872          & 0.2614          & \textbf{0.3429} \\
\bottomrule
\end{tabular}
\caption{ The comparison result of 3D object detection on nuScenes validation set.
}
\label{tab:supp_detection}
\end{table*}

\begin{table*}[h]
\setlength{\abovecaptionskip}{3pt} 
\setlength{\belowcaptionskip}{0pt}
\setlength{\tabcolsep}{2mm}
\centering
\renewcommand{\arraystretch}{1.3}
\small
\begin{tabular}{l|ccccccc}
\toprule
Method & Drivable $\uparrow$ & Ped\_crossing $\uparrow$ & Walkway $\uparrow$ & Stopline $\uparrow$ & Carpark\_area $\uparrow$ & Divider $\uparrow$ & Mean $\uparrow$   \\
\midrule
\rowcolor{gray!30} Oracle~\cite{liu2023bevfusion}       & 0.8173   & 0.5454       & 0.5846  & 0.4735   & 0.5400         & 0.4649  & 0.5709 \\
MagicDrive-V~\cite{gao2024magicdrive3d} & 0.3858   & 0.0751       & 0.1546  & 0.1204   & 0.1659       & 0.1245  & 0.1710  \\
\rowcolor{dcolor} CogDriving (Ours)        & 0.7192   & 0.4663       & 0.5000     & 0.4126   & 0.4701       & 0.4014  & 0.4953 \\
\bottomrule
\end{tabular}
\caption{The comparison result of BEV segmentation on nuScenes validation set
}
\label{tab:supp_segmentation}
\end{table*}

\begin{figure*}[h]
    \centering
    \includegraphics[width=0.99\linewidth]{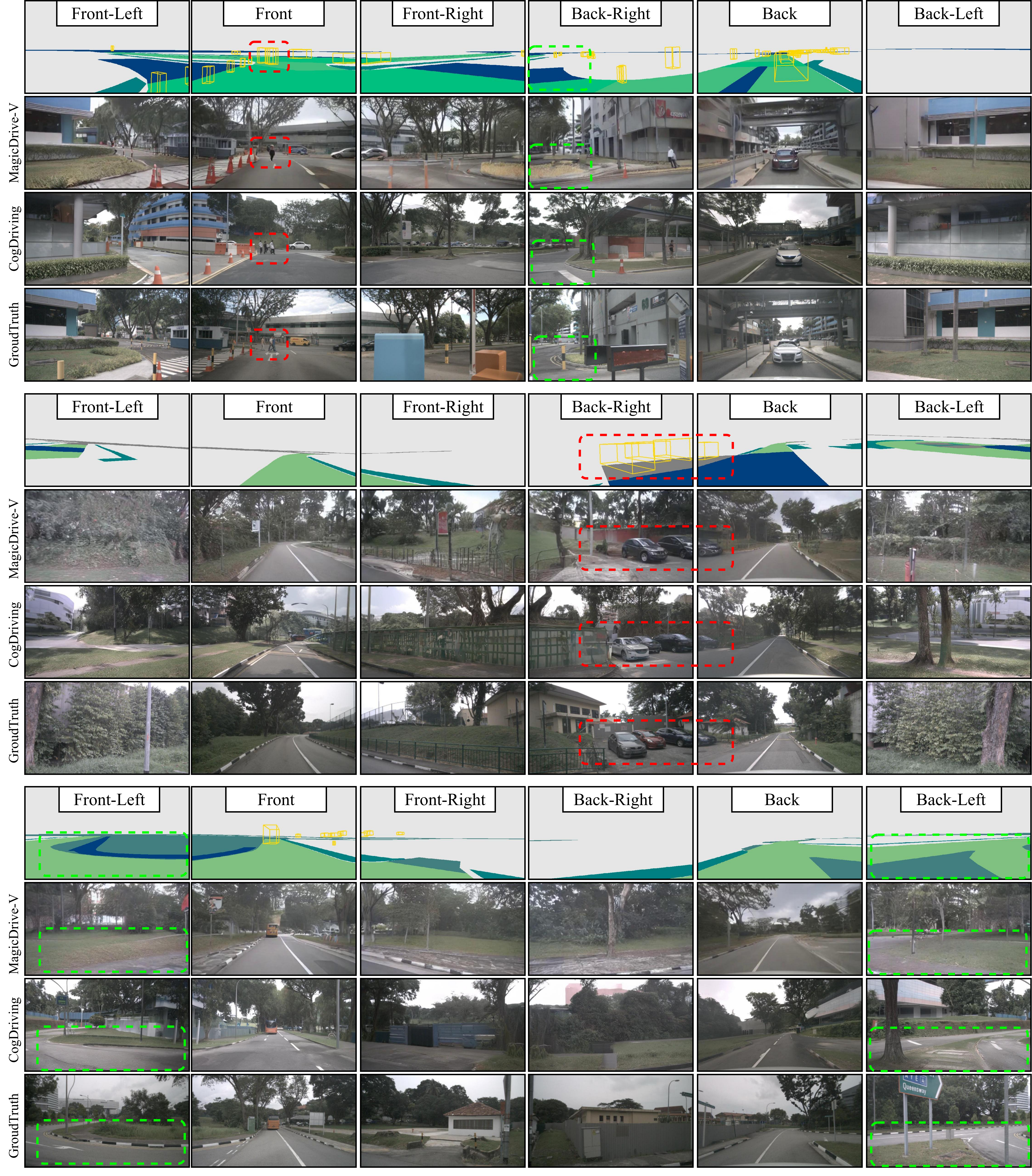}
    \caption{Qualitative comparison for controllability. 
    The red boxes indicate the regions where objects should be generated based on the condition of 3D bounding boxes. 
    The green boxes indicate the regions where corresponding roads should be generated based on the BEV road map.
    }
    \label{fig:supp_1}
\end{figure*}

\section{Qualitative Analysis: Controllability}
\label{sec:vis_results}
As a supplement to Sec.5.3 of the main manuscript, we provide additional visualization results for qualitative analysis in Fig~\ref{fig:supp_1}.
Since MagicDrive-V~\cite{gao2024magicdrive3d} is the only multi-view video generation method that releases model weights, we use the generation results from MagicDrive-V for comparison.
As shown in Fig~\ref{fig:supp_1}, compared with MagicDrive-V, our CogDriving is capable of generating a diverse range of objects, such as pedestrians and vehicles, as well as more realistic road areas.

\section{Training Details}
\label{sec:train_detail}

The training split of the NuScenes~\cite{caesar2020nuscenes} dataset contains 700 video clips, each lasting about 20 seconds, with a spatial resolution of 900$\times$1600.
We preprocess these videos by segmenting them into clips of 16 frames.
%
The resolution of these clips is resized to 405$\times$720, followed by zero-padding along the shorter edges to 480$\times$720 to better align with the pre-trained model weights.
For BEV sequence conditions, we apply the same preprocessing steps, including resizing to 405$\times$720 and zero-padding to 480$\times$720, ensuring alignment with the frame content.
During the second stage of training, we fine-tune the model using a LoRA rank of 256, resulting in a total of 172.18M trainable parameters.